\documentclass[11pt]{article}

\usepackage[preprint]{acl}

\usepackage{times}
\usepackage{latexsym}

\usepackage[T1]{fontenc}
\usepackage[utf8]{inputenc}
\usepackage{microtype}

\usepackage{graphicx}
\usepackage{amsmath, amssymb, amsfonts}
\usepackage{booktabs}
\usepackage{pifont}
\usepackage{listings}

\makeatletter
\renewcommand{\@makefntext}[1]{\noindent\@makefnmark\ #1}
\renewenvironment{abstract}%
  {\begin{center}\large\textbf{\abstractname}\end{center}%
    \begin{list}{}%
      {\setlength{\rightmargin}{0cm}%
        \setlength{\leftmargin}{0cm}}%
      \item[]\ignorespaces%
      \@setsize\normalsize{12pt}\xpt\@xpt
  }%
  {\unskip\end{list}}
\makeatother

\lstdefinestyle{autovend-prompt}{
    basicstyle=\ttfamily\footnotesize,
    columns=fullflexible,
    keepspaces=true,
    breaklines=true,
    breakatwhitespace=false,
    breakautoindent=false,
    breakindent=0pt,
    showstringspaces=false,
    frame=single,
    backgroundcolor=\color{black!5},
    rulecolor=\color{black!25},
    framerule=0.4pt,
    framesep=4pt,
    xleftmargin=4.4pt,
    xrightmargin=4.4pt,
    literate={—}{{\textemdash}}1
}
\newcommand{\cmark}{\textcolor{green!55!black}{\ding{51}}}
\newcommand{\xmark}{\textcolor{red!80!black}{\ding{55}}}
\newcommand{\hdr}[1]{\textbf{\begin{tabular}{@{}c@{}}#1\end{tabular}}}
\newcommand{\Description}[1]{}

\title{AutoVend: Benchmarking LLM Agents on Long-Horizon Autonomous Vending Operations}

\author{
  \textbf{Qiming Shi\textsuperscript{1,2}\thanks{\raggedright Equal contribution. Work done during internship at StepFun}},
  \textbf{Yikun Li\textsuperscript{1,3}\footnotemark[1]},
  \textbf{Beidi Luan\textsuperscript{1}},
  \textbf{Rui Sun\textsuperscript{1}},
\\
  \textbf{Yan Ma\textsuperscript{1}},
  \textbf{Jing Li\textsuperscript{1}},
  \textbf{Daxin Jiang\textsuperscript{1}},
  \textbf{Zuo Bai\textsuperscript{1,4}\thanks{\raggedright Corresponding author: Zuo Bai (\texttt{baizuo@stepfun.com}).}}
\\
\\
  \textsuperscript{1}StepFun,
  \textsuperscript{2}Zhejiang University,
  \textsuperscript{3}Xi'an Jiaotong-Liverpool University,
  \textsuperscript{4}FinStep
\\
  {\small \texttt{\{baizuo\}@stepfun.com}}
}

\begin{document}
\maketitle

\begin{abstract}
Large language model (LLM) agents are extending agent-mediated commerce on the Web from shopping for consumers to operating businesses for sellers.
In online-to-offline (O2O) commerce, decisions are issued through Web-based platforms but realized through costly physical execution.
Evaluating such agents requires going beyond individual transactions to ask whether decisions made online turn into sustained economic returns.
We introduce AutoVend, a benchmark in which one LLM agent manages 20 vending machines across five site types through a Web-based operating interface for 180 simulated days.
Its data-grounded simulator combines a shared warehouse, procurement lead times, paid replenishment, inventory expiration, and recurring operating costs.
We evaluate five LLM agents against a rule-based inventory-control baseline and diagnose their site coordination, execution, and performance over time.
Four of five evaluated LLM agents fall short of this baseline, and returns per dollar of inference cost rank models differently from returns alone.
Operational analyses show that gains in individual metrics do not consistently yield stronger economic growth, as inventory and execution costs erode these gains.
Trajectory analysis further reveals difficulties in grounding feasible actions, managing inventory commitments, and converting feedback into effective policy revisions.
\end{abstract}

\section{Introduction}
Agent-mediated commerce is emerging as a Web application domain for large language model (LLM) agents, with buyer-side research spanning product discovery~\cite{yao2022webshop}, personalized shopping~\cite{kim2026agenticshop}, and customer support~\cite{wang2025ecombench}.
Seller-side autonomy requires repeated decisions about assortment, pricing, procurement, and inventory whose consequences extend beyond an individual interaction~\cite{merchantbench,ecommercebench}.
In online-to-offline (O2O) commerce, decisions are made on Web-based management platforms, while their economic returns depend on the costs and constraints of physical execution.
For agents managing distributed retail through such platforms, this requires coordinating digital decisions with physical execution over time.
Smart vending is a compact instance, as connected machines are increasingly run through cloud management platforms~\cite{fbi2026vending} while every restock requires a paid site visit.
It also retains the dependencies among assortment, inventory, execution, and cash management.
For Web economics, AutoVend thus provides a controlled testbed for seller-side agent-mediated commerce that measures both the returns an agent generates through a Web-based interface and the inference cost of generating them.

\begin{figure}[t]
\centering
\includegraphics[width=\columnwidth]{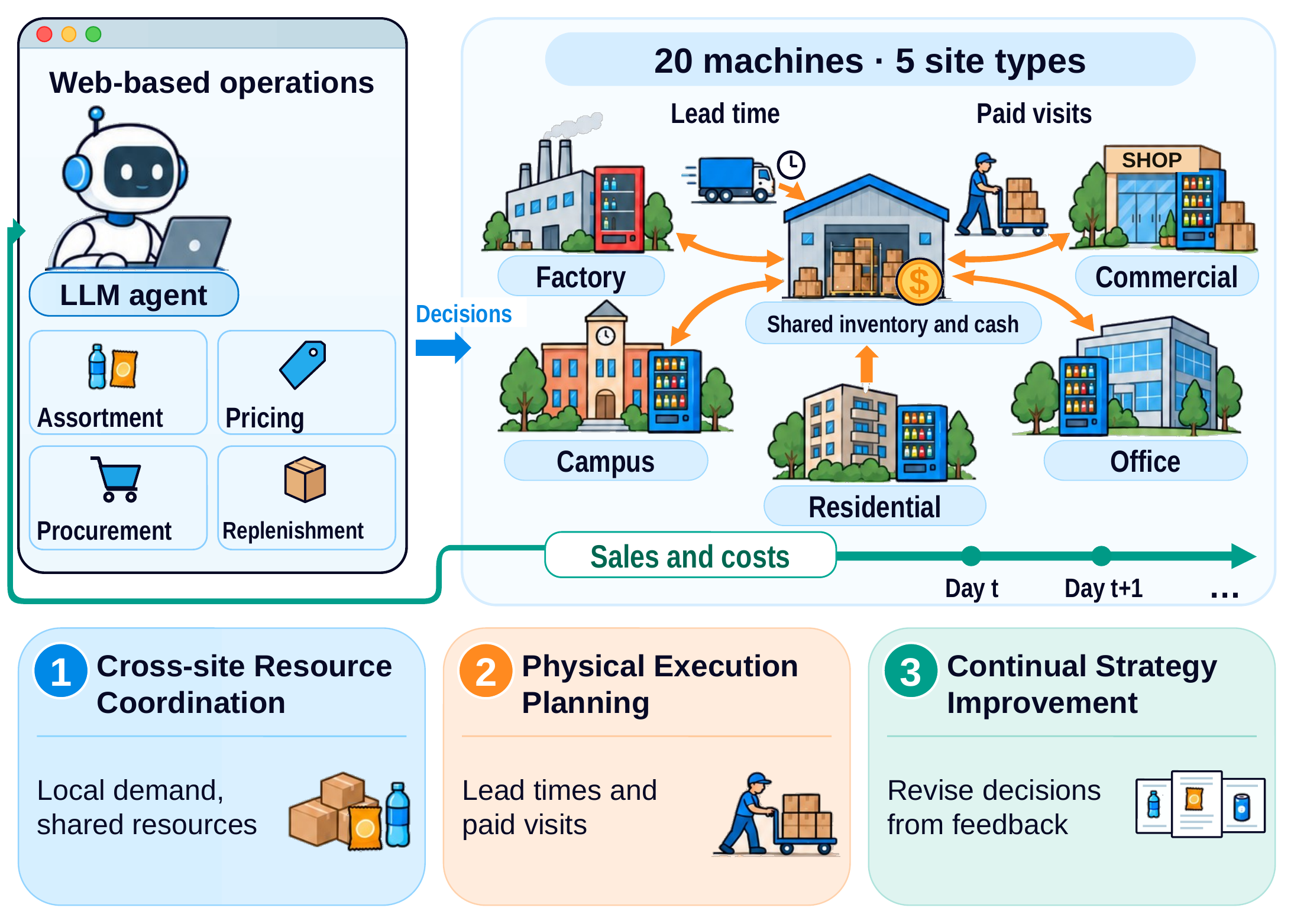}
\caption{AutoVend connects Web-based agent decisions to distributed physical retail through shared inventory, procurement lead times, and paid on-site replenishment.}
\Description{An LLM agent manages vending machines across five site types through a Web-based interface, with shared inventory and cash, procurement lead times, paid visits, and feedback from sales and costs.
The diagram highlights cross-site resource coordination, physical execution planning, and continual strategy improvement.}
\label{fig:intro}
\end{figure}

Sustaining performance in this setting requires three interdependent capabilities.
\textbf{(1) Cross-site Resource Coordination.}
The agent must match heterogeneous local demand while allocating shared resources and monitoring conditions across the network~\cite{eppen1979centralization}.
An allocation to one site changes the resources available to others, so local decisions must account for consequences across the network.
\textbf{(2) Physical Execution Planning.}
Stock in a shared warehouse cannot meet local demand until it is placed on machine shelves.
Procurement requires advance ordering to account for delivery lead times, while on-site replenishment incurs visit costs.
The agent must coordinate advance orders and site visits to maintain product availability while controlling execution expenses, storage costs, and expiration losses~\cite{coelho2014inventoryrouting,nahmias1982perishable}.
\textbf{(3) Continual Strategy Improvement.}
Demand is initially unknown, and observed sales depend on the agent's own assortment, prices, and stock availability.
The agent must use accumulated feedback to refine its demand estimates and revise its decision rules so that subsequent decisions become more effective~\cite{retailbench,merchantbench}.

Existing benchmarks provide important foundations for evaluating long-horizon business agents.
Vending-Bench studies the operation of a single vending machine and simulates physical tasks through a separate execution agent~\cite{vendingbench}.
RetailBench evaluates supermarket operation with inventory aging, delayed deliveries, and cash-flow constraints~\cite{retailbench}.
MerchantBench examines seller-side e-commerce under mixed-latency feedback~\cite{merchantbench}, while E-Commerce Bench extends business evaluation to multiple online stores with pooled inventory, fulfillment costs, and delayed settlements~\cite{ecommercebench}.
These evaluations do not establish how effectively agents can coordinate shared inventory across heterogeneous physical retail sites when on-site replenishment itself incurs costs and delays.
Assessing this capability requires an environment that brings these constraints together and exposes their cumulative economic consequences.

We introduce \textbf{AutoVend}, a benchmark for evaluating LLM agents on long-horizon autonomous vending operations.
A single agent manages 20 vending machines spanning factory, campus, residential, commercial, and office locations, with heterogeneous demand and a shared warehouse.
The agent selects products, sets prices, places procurement orders, and arranges replenishment through tool calls that mirror such a platform.
The simulator models procurement lead times, paid on-site stocking, product expiration, and storage charges on warehouse stock.
These mechanisms connect decisions made through the interface to inventory availability and subsequent economic outcomes.

AutoVend combines terminal net assets with operational metrics and trajectory diagnostics.
We examine site coordination, the cost and availability of paid replenishment, and changes in operating performance over time.
Our contributions are threefold.
\begin{itemize}
    \item We introduce AutoVend, a benchmark for long-horizon seller-side agent-mediated commerce, with a unified task, operating interface, and evaluation protocol for managing a distributed retail network.
    \item We develop a simulator grounded in real vending records that models heterogeneous demand, procurement lead times, inventory aging, and paid site visits, and validate its demand behavior against observed sales patterns.
    \item We evaluate five LLM agents and find that four of five fall short of a rule-based inventory-control baseline.
    Operational diagnostics and trajectory analyses reveal difficulties in coordinating inventory commitments, execution costs, and policy revision.
\end{itemize}

\section{Related Work}

\textbf{Agent Evaluation in Commerce.}
Commerce is a prominent domain for evaluating LLM agents on the Web, with benchmarks for online shopping~\cite{yao2022webshop,peeters2026webmall,kim2026agenticshop}, customer service~\cite{yao2025taubench,wang2025ecombench}, and multi-agent markets~\cite{bansal2025magentic,zheng2026marketbench}.
Their evaluations typically end with a completed task, dialogue, or transaction, whereas a seller's decisions affect inventory and cash long after they are made.
Seller-side benchmarks instead evaluate the sustained operation of a business mainly by its accumulated economic outcome.
MerchantBench and StoreBench each cover a single online store~\cite{merchantbench,storebench}, Business Arena a single cross-border shop~\cite{businessarena}, and E-Commerce Bench up to four online stores that share a warehouse~\cite{ecommercebench}.
In physical retail, RetailBench covers a single supermarket~\cite{retailbench} and Vending-Bench a single vending machine~\cite{vendingbench}, a setting that Vending-Bench 2 extends to a simulated year and its Arena to several competing agents with one machine each~\cite{andonlabs2025vendingbench2}.
AutoVend instead provides a controlled setting in which one agent operates 20 heterogeneous sites from a shared warehouse and every on-site action requires a paid visit (Table~\ref{tab:related}).

\begin{table*}[t]
\centering
\caption{Comparison with long-horizon business operation benchmarks.
Op.\ Units counts the operating units managed by one agent.
Diverse Types marks whether these units differ in type.
Real Data marks whether the environment is grounded in real commercial datasets.
Physical Ops.\ marks whether actions require physical execution.
Open Source marks public release, upon publication for AutoVend.}
\label{tab:related}
\small
\setlength{\tabcolsep}{6pt}
\begin{tabular}{@{}lcccccc@{}}
\toprule
\textbf{Benchmark} & \textbf{Domain} & \hdr{Op.\\Units} & \hdr{Diverse\\Types} & \hdr{Real\\Data} & \hdr{Physical\\Ops.} & \hdr{Open\\Source} \\
\midrule
RetailBench~\cite{retailbench} & Supermarket & 1 & \xmark & \cmark & \xmark & \cmark \\
MerchantBench~\cite{merchantbench} & E-commerce & 1 & \xmark & \cmark & \xmark & \cmark \\
StoreBench~\cite{storebench} & E-commerce & 1 & \xmark & \xmark & \xmark & \xmark \\
E-Commerce Bench~\cite{ecommercebench} & E-commerce & $\leq$4 & \cmark & \cmark & \xmark & \cmark \\
Vending-Bench~\cite{vendingbench} & Vending & 1 & \xmark & \xmark & \cmark & \xmark \\
\midrule
\textbf{AutoVend (ours)} & Vending & 20 & \cmark & \cmark & \cmark & \cmark \\
\bottomrule
\end{tabular}
\end{table*}

\textbf{Long-Horizon Agent Evaluation.}
Long-horizon benchmarks commonly lengthen the work toward a defined goal, as in workflows on websites and computers~\cite{zhou2024webarena,yuan2026osworld2}, tasks in a simulated company~\cite{xu2025theagentcompany}, and exploration of partially observable environments~\cite{luo2026ultrahorizon}.
Sustained-operation benchmarks instead fix a span of simulated calendar time in which costs accrue and the environment evolves whether or not the agent acts, so performance is an accumulated economic result rather than goal completion~\cite{vendingbench,he2026ycbench}.
Because terminal outcomes alone reveal little about why agents succeed or fail, evaluations add progress rates~\cite{ma2024agentboard}, capability profiles across operating dimensions~\cite{ecommercebench}, and attribution of gains and losses to actions~\cite{businessarena}.
Continual-learning benchmarks further ask whether agents turn accumulated experience into reusable capabilities~\cite{zheng2025lifelongagentbench,cheng2026ahabench}.
AutoVend ties its operational dimensions to individual sites and on-site execution and examines how performance changes across operating scales and over time.

\section{The AutoVend Benchmark}
\label{sec:autovend}
\begin{figure*}[t]
\centering
\includegraphics[width=\textwidth]{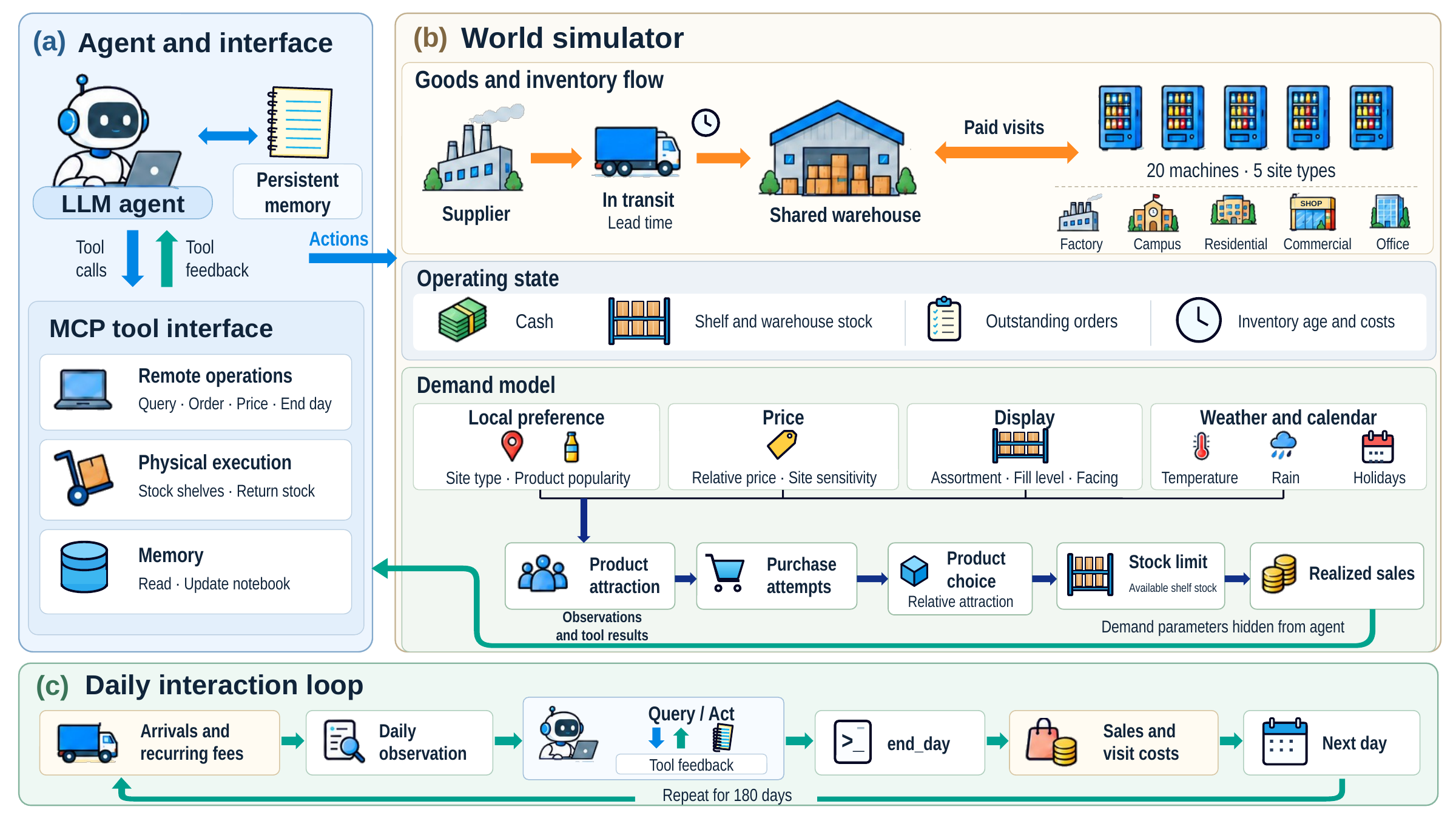}
\caption{Overview of AutoVend.
(a) Agent and interface, with MCP tools for remote operations, physical execution, and persistent memory.
(b) World simulator, connecting a shared warehouse to 20 machines across five site types and generating sales through a data-grounded demand model with hidden parameters.
(c) Daily interaction loop, from arrivals and recurring fees through agent queries and actions to sales and visit-cost settlement, followed by advancement to the next day.}
\Description{Diagram of the AutoVend environment showing the agent, its tool interface, the shared warehouse, the vending machines, and the demand model.}
\label{fig:overview}
\end{figure*}

AutoVend places an LLM agent in charge of a network of vending machines over a long operating horizon (Figure~\ref{fig:overview}).
The agent sets assortment and prices, orders stock, and arranges on-site replenishment, while customer demand is grounded in real operating data and hidden from the agent.

\subsection{Task Formulation}
\label{sec:task}
An agent operates 20 vending machines that share one warehouse and draw on a catalog of 1,372 products over 180 simulated days, starting with 40,000~CNY in cash.
We formulate the task as a partially observable Markov decision process (POMDP)~\cite{kaelbling1998planning}.
The underlying state $s_t$ comprises the simulation date, cash, inventory, outstanding orders, machine assortments and prices, and hidden demand conditions.
An action $a_t\in\mathcal A$ denotes a tool invocation.
When the agent ends the day's actions, the simulator settles sales and costs and advances to the next day.
The observation kernel $Z(o_{t+1}\mid s_{t+1},a_t)$ exposes daily operating summaries and tool results, while demand parameters and unmet demand remain hidden.

The objective is to maximize terminal net assets,
\begin{equation}
\mathrm{NA}_T = B_T + \sum_{k} c_k h_{k,T},
\label{eq:na}
\end{equation}
where $T$ is the terminal interaction step at the end of the operating episode, $B_t$ is cash, $c_k$ is the unit cost of product $k$, and $h_{k,t}$ counts the unexpired units of $k$ on shelves, in the warehouse, or in transit.
Buying stock converts cash into inventory at cost, so net assets change through sales, operating costs, fines, and inventory write-offs from expiration or disposal.
With underlying demand and its response parameters hidden, the agent can refine its assortment, pricing, and replenishment strategy using feedback accumulated during operation.

\begin{figure*}[t]
\centering
\includegraphics[width=\textwidth,trim=0 6mm 0 0,clip]{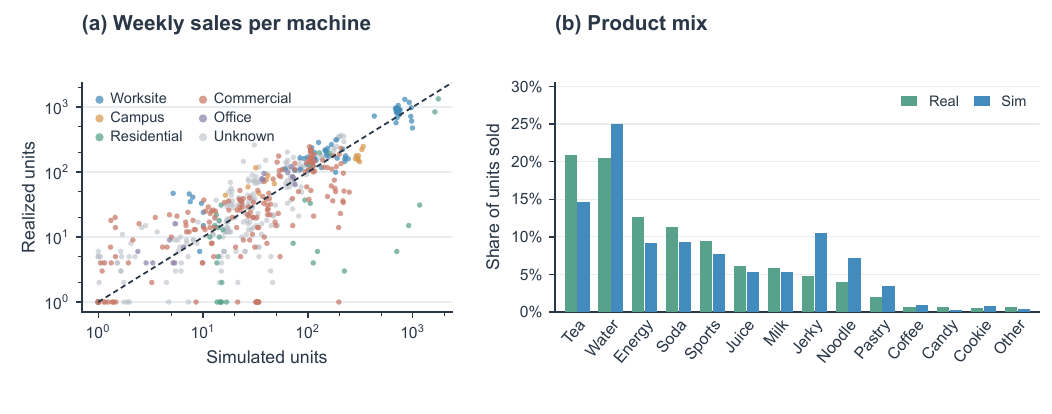}
\caption{Held-out evaluation of the demand model on 67 machines over an eight-week test window, averaged over 100 seeds.
(a) Simulated versus realized weekly sales per machine.
(b) Subcategory shares of realized and simulated sales.
Subcategories accounting for less than 0.5\% of realized sales are pooled into Other, while TVD uses all original subcategories.}
\Description{Two panels comparing simulated and realized machine-week sales and subcategory sales shares.}
\label{fig:validation}
\end{figure*}

\subsection{World Simulator}
\label{sec:simulator}
\textbf{Network and operations.}
The network comprises six worksite, two campus, four residential, six commercial, and two office machines, with differences in product preference, price sensitivity, and holiday response.
Each machine holds at most 200 units and is replenished from one shared warehouse.
Procurement is paid at placement.
Orders arrive in the warehouse three days after placement.
Staff are available for on-site visits on working days, and multiple visits can be made within a day.
Replenishment and removal of shelf stock require physical execution through an on-site visit, which costs 35~CNY per machine per day regardless of the number of operations.

\textbf{Inventory aging and costs.}
Batches expire 30 or 90 days after warehouse arrival.
Selling expired products incurs a fine of 1,000~CNY per unit sold.
Warehouse rent is 0.03~CNY per unit per day, and machine rent is 4~CNY per machine per day.
The agent must balance stock availability against storage costs and expiration risk, since excess inventory incurs recurring charges and may expire before it is sold.

\textbf{Customer demand.}
Demand is generated separately for each machine, with temporal dependence introduced by persistent daily shocks.
The attraction of product $k$ on machine $m$ is
\begin{equation}
\begin{aligned}
w_{m,k,t} = {}& b_{m,k}\, F_{m,t}^{\eta} \min\Big(1,\frac{q_{m,k,t}}{\phi}\Big) \\
& \times \Big(\frac{p_{m,k,t}}{\bar{p}_k}\Big)^{-\varepsilon_{\ell(m)}} E_{m,k,t},
\end{aligned}
\label{eq:attraction}
\end{equation}
where $b_{m,k}$ is a base attraction, $F_{m,t}$ the filled fraction of capacity, $q_{m,k,t}$ the shelf stock, $p_{m,k,t}$ the price, $\bar{p}_k$ the reference price, and $\varepsilon_{\ell(m)}$ the price elasticity of the site type.
Fuller machines attract more customers~\cite{balakrishnan2004stack}, and an item displayed with fewer than $\phi=5$ units loses attraction proportionally.
The multiplier $E_{m,k,t}$ captures temperature, rain, and holiday effects.
Purchases follow
\begin{equation}
\begin{gathered}
N_{m,t} \sim \mathrm{Poisson}\Big(\xi_{m,t}\frac{U_m W_{m,t}}{U_m+W_{m,t}}\Big), \\
W_{m,t}=\sum_{k} w_{m,k,t},
\end{gathered}
\label{eq:purchases}
\end{equation}
Here, $N_{m,t}$ counts purchase attempts, and $U_m$ is the machine-specific weight assigned to the option of buying nothing.
Each purchase attempt selects product $k$ with probability $w_{m,k,t}/W_{m,t}$.
This implements a multinomial logit model with a no-purchase option~\cite{talluri2004revenue}.
The random multiplier $\xi_{m,t}$ averages one and is positively correlated across days, allowing demand fluctuations to persist and increasing variability beyond Poisson sampling.
Sales equal the minimum of sampled demand and shelf stock, and unmet demand is lost.

\subsection{Real-World Data Grounding}
\label{sec:grounding}
\textbf{Data.}
We build the simulator from six months of order-level sales records covering 99 machines operated by a commercial smart vending company (Appendix~\ref{app:ethics}), totaling 220,128 units sold.
We select 20 machines across five site types to construct the simulated network operated by the agent.
The catalog keeps 1,372 products in 18 subcategories, covering 89.3\% of units sold.

\textbf{Parameter calibration.}
The base attraction in Eq.~\eqref{eq:attraction} factorizes as
\begin{equation}
b_{m,k} = \kappa_m\, \pi_{\ell(m),g_1(k)}\, \gamma_{g_2(k)}\, \omega_k,
\end{equation}
where $g_1(k)$ and $g_2(k)$ are the category and subcategory of product $k$.
The product popularity $\omega_k$ is the number of units sold per unit of exposure, computed from sales records and shrunk toward its subcategory mean by empirical Bayes.
Exposure counts a machine's days of offering the product weighted by its sales level.
The site preference $\pi_{\ell,g_1}$ compares the drink and snack mix of labeled machines with the mix expected from the products they actually sold, so that stocking choices are not mistaken for site effects.
The coefficients $\gamma_{g_2}$ align predicted subcategory shares on real shelves with realized shares.
The scale $\kappa_m$ sets the expected units of a typical real shelf at reference prices and neutral weather equal to the machine's observed mean daily units.

The no-purchase weight is $U_m = u\,R_m$ with $u = x e/(1-e)$, where $x$ is the geometric mean ratio between the attraction of real shelves and that of a fully stocked reference shelf, $e$ is the elasticity of daily units with respect to shelf attraction estimated within machines, and $R_m$ is the attraction of the reference shelf.
The reference shelf sorts products by $b_{m,k}$ and places $\phi$ units of each until capacity is full.
With $x=0.0638$ and $e=0.318$, a typical real shelf sells the machine's observed mean daily units, and a fully stocked reference shelf sells about $1.42$ times that amount.
The shock is $\xi_{m,t}=\exp(\sigma u_{m,t}-\sigma^2/2)$ with $u_{m,t}=\sum_{j=0}^{5}\rho^{j} z_{m,t-j}/\sqrt{\sum_j \rho^{2j}}$, where $z_{m,t}$ are independent standard normal draws and $\sigma^2=\ln(1+1/r)$.
The dispersion $r=2.74$ is the median overdispersion $m^2/(v-m)$ of daily units after removing weekday patterns, and $\rho=0.21$ is the lag-one residual autocorrelation.
The nested structure of the choice model keeps one nest per subcategory with similarity parameter $\lambda=1$, at which the model reduces to the multinomial logit form used above.

Margins, shelf lives, price elasticities, environmental responses, and operating costs are benchmark settings.
We set the fill exponent to $\eta=0.5$, the facing threshold to $\phi=5$ units, and site-specific price elasticities between 2.4 and 3.5.
Unit costs use assumed gross margins of 30--40\% for drinks and 50--60\% for snacks.

\textbf{Validation.}
We re-estimate all data-derived parameters from records preceding an eight-week test window and evaluate the demand model on 67 machines.
We reconstruct daily shelves from inventory records for 67 machines.
We replay these shelves at reference prices and under neutral weather, comparing simulated with realized sales over 100 seeds (Figure~\ref{fig:validation}).

Let $v_t$ denote the realized sales for machine-week observation $t$.
The weighted absolute percentage error is
\begin{equation}
\mathrm{WAPE} = \frac{\sum_{t} \lvert v_t - \hat{v}_t \rvert}{\sum_{t} v_t},
\end{equation}
where $\hat{v}_t$ is the simulated counterpart.
Subcategory fidelity is measured by the total variation distance between simulated and realized subcategory shares, and product ranking by the Spearman correlation between simulated and realized product sales within a machine.
The persistence reference sets $\hat{v}_t$ to each machine's mean weekly sales over the preceding four weeks, and the naive reference keeps each machine's earlier sales level and spreads it uniformly over the products on its shelf.
Two ablations show which components matter, removing the overdispersed shock and replacing the estimated popularity with a uniform weight.

Total volume is reproduced within 0.7\%, with a machine-week weighted absolute percentage error (WAPE) of 0.473 against 0.606 for a naive reference and 0.561 for a persistence reference that carries forward the preceding four weeks of real sales.
The total variation distance (TVD) between simulated and realized subcategory shares is 0.153, compared with 0.206 for the naive reference.
Within machines, the median Spearman correlation of product sales is 0.577.
A uniform product popularity raises the TVD to 0.238, showing that estimated product popularity improves mix fidelity.
Price and environmental responses remain unvalidated by design.

\subsection{Agent Interface}
\label{sec:interface}
\textbf{Interaction loop.}
AutoVend exposes the remote operations available to a networked-vending operator as tools over the Model Context Protocol (MCP)~\cite{anthropic2024mcp}, with physical tools standing in for dispatching staff on site.
Each day begins with due order arrivals and charges for warehouse rent and fixed operating fees, followed by a daily observation for the agent.
During the action window, the agent queries operating records and executes actions, receiving query results or execution feedback after each tool call.
Valid actions update the operating state immediately.
When the agent closes the action window, the simulator realizes demand using the resulting shelf inventory and prices, settles sales and on-site visit costs, and proceeds to the next day.

\textbf{Prompts.}
A fixed system prompt specifies the objective, starting conditions, goods flow, operating costs, inventory and pricing constraints, and qualitative demand mechanisms without prescribing an operating strategy.
A daily user prompt updates the date and supplies a compact report of cash and net assets, recent profit and loss, warehouse and in-transit stock, upcoming holidays, and each machine's inventory, sales, and expiry status.
Product-level details are retrieved through queries rather than included in the daily prompt.
Appendices~\ref{app:prompts} and~\ref{app:tools} provide the complete prompts and tool specifications.

\textbf{Tools and memory.}
AutoVend provides 18 tools organized into three functional groups covering remote operations, physical execution, and persistent memory.
\begin{itemize}
    \item \textbf{Remote operational tools.}
    These support procurement, machine-specific pricing, warehouse disposal, and closing the daily action window.
    Queries provide product attributes and costs, warehouse and shelf inventory, product-level sales within and across machines, and financial records.
    Read-only SQL supports custom analysis of operating records, while calendar and weather queries provide published holidays and near-term weather information.

    \item \textbf{Physical execution tools.}
    These transfer warehouse stock onto machine shelves and return specified quantities or all remaining units of a product to the warehouse.
    Execution requires paid on-site visits, charged once per machine per day regardless of the number of operations.

    \item \textbf{Memory tools.}
    These allow the agent to read and overwrite a persistent notebook containing selected observations, plans, and findings.
    Conversation history continues across days, while the notebook survives context compaction.
\end{itemize}

\section{Experiments}
\label{sec:experiments}

\subsection{Experimental Setup}
\textbf{Baseline.}
We evaluate Claude Sonnet 5.5~\cite{claudesonnet55}, GPT-6.1 Sol~\cite{gpt61sol}, DeepSeek V4.1 Flash~\cite{deepseekv41flash}, Kimi K3~\cite{kimik3}, and GLM-5.3-Flash~\cite{glm53flash}.
We use the Pi harness~\cite{zechner2026pi} and configure each model with its highest supported reasoning effort.
Each agent operates the same 20-machine network for 180 simulated days, starting with 40,000~CNY and using identical prompts, tools, and daily interaction budgets.
We report means and the standard deviation of terminal net assets over three runs per model, all in one frozen world with the same random seed.
Variation across LLM runs thus reflects stochastic agent decisions and the demand draws they induce.
The deterministic rule-based baseline attains $208.73 \pm 2.26$ thousand CNY (CV 1.1\%) across seeds 1--3, and we report seed 1.

The rule-based baseline combines the newsvendor critical ratio~\cite{arrow1951optimal} with periodic order-up-to replenishment~\cite{silver2008reorder}.
It sets inventory targets from rolling 30-day demand and variability estimates, balancing lost sales against holding and spoilage costs, and replenishes sites weekly.
Procurement accounts for delivery lead times, available stock, and outstanding orders.
Assortments follow observed sales, while prices are adjusted through local gross-profit trials.
All decisions use observable information.

\textbf{Metrics.}
Our primary outcome, \textbf{Net Assets}, is terminal cash plus unexpired inventory valued at cost across shelves, the warehouse, and transit.
\textbf{Assortment} compares expected gross profit with a heuristic reference mix under common prices, display volume, and environmental conditions.
\textbf{Waste Rate} is inventory loss divided by valid sales cost plus loss, counting expiry and early disposal once.
\textbf{Pricing Regret} is the expected gross-profit shortfall from a price-search reference at a fixed site, assortment, and inventory.

\textbf{Shelf Utilization} measures the mean fraction of capacity occupied by unexpired stock before sales across sites and days.
We also report the share of site-days with empty shelves.
\textbf{Capital Efficiency} is mean daily gross profit divided by mean inventory value at cost.
\textbf{Tool Calls} counts tool invocations over the episode.
\textbf{API Efficiency} averages per-run net asset growth per USD of estimated API expenditure, using standard reference rates for recorded input, output, cache, and compaction usage.

\subsection{Main Results}
\label{sec:main-results}

\begin{table*}[t]
\centering
\caption{Main experimental results on AutoVend.
LLM results are averaged over three repeated runs, and Rule-based reports seed 1.
Net Assets and Std are in thousands of CNY.
Waste Rate and Shelf Utilization are percentages, Pricing Regret is in CNY per machine per day, and API Efficiency is net-asset growth in CNY per USD of API cost.}
\label{tab:main-results}
\small
\setlength{\tabcolsep}{2pt}
\begin{tabular*}{\textwidth}{@{\extracolsep{\fill}}lcccccccc@{}}
\toprule
\textbf{Model} & \hdr{Net\\Assets\ $\uparrow$} & \textbf{Std} & \hdr{Assortment\ $\uparrow$} & \hdr{Waste\\Rate\ $\downarrow$} & \hdr{Pricing\\Regret\ $\downarrow$} & \hdr{Shelf\\Utilization} & \hdr{Tool\\Calls} & \hdr{API\\Efficiency\ $\uparrow$} \\
\midrule
GPT-6.1 Sol & \textbf{309.35} & 34.09 & 0.992 & 0.60 & 30.18 & 88.91 & 3,950 & 1,117 \\
GLM-5.3-Flash & 148.43 & 16.70 & 0.817 & 0.27 & 2.45 & 60.65 & 670 & 7,023 \\
DeepSeek V4.1 Flash & 133.92 & 27.20 & 0.622 & 1.85 & 4.22 & 82.40 & 2,118 & 31,010 \\
Claude Sonnet 5.5 & 103.04 & 11.33 & 0.500 & 0.16 & 1.32 & 32.48 & 1,013 & 3,235 \\
Kimi K3 & 91.52 & 41.75 & 0.661 & 0.60 & 3.29 & 42.29 & 517 & 1,920 \\
Rule-based & 210.97 & 0.00 & 1.034 & 0.17 & 4.32 & 64.65 & 1,624 & N/A \\
\bottomrule
\end{tabular*}
\end{table*}

\begin{figure*}[t]
\centering
\includegraphics[width=\textwidth]{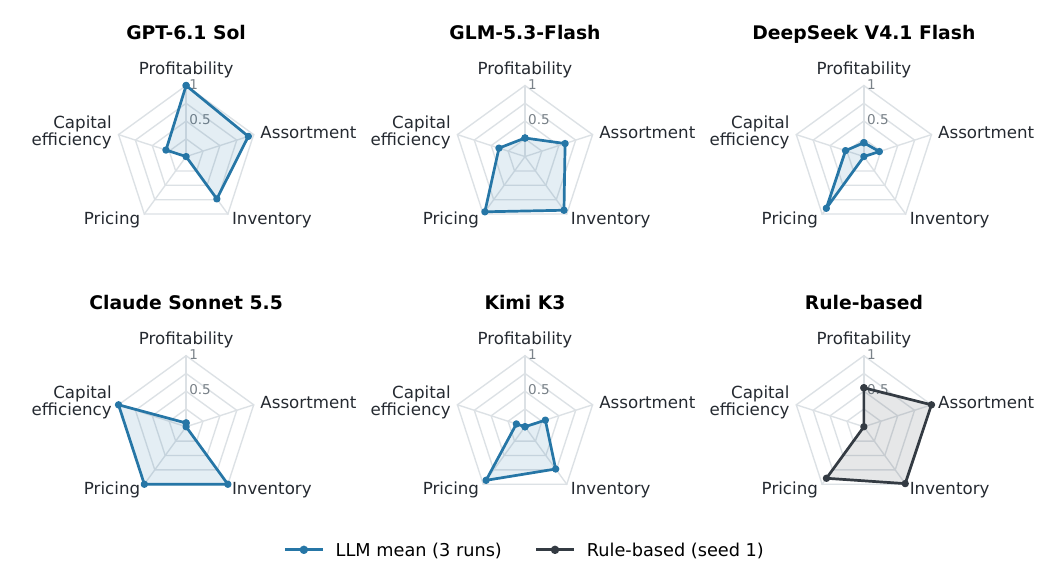}
\caption{Whole-episode operational profiles over 180 days.
Each panel shows the mean operational profile of an LLM or the rule-based baseline.
Profitability, Assortment, Inventory, Pricing, and Capital Efficiency correspond to terminal net assets, assortment quality, reversed waste rate, reversed pricing regret, and daily gross profit per unit of inventory capital, respectively.
Each axis is min-max normalized across the five model means and the baseline, with larger values indicating better relative performance.}
\label{fig:operational-profiles}
\end{figure*}

\textbf{Overall Performance.}
Long-horizon business management remains challenging, with four of five evaluated models underperforming the rule-based baseline in Table~\ref{tab:main-results}.
GPT-6.1 Sol is the sole exception, while all five models obtain lower assortment scores than the baseline.
All three GPT executions exceed the baseline, whereas none of the other 12 does, showing that most models' weak averages cannot be explained by a few isolated failures.

\textbf{Variability.}
Weak average performance is compounded by inconsistent execution.
Kimi K3 exhibits a coefficient of variation of 45.6\% across repeated runs.
GPT-6.1 Sol, GLM-5.3-Flash, and Claude Sonnet 5.5 instead exhibit similar relative dispersion of around 11\%, despite their substantially different mean returns.
This contrast shows that consistency and economic effectiveness are distinct aspects of agent capability.

\textbf{API Efficiency.}
Operating returns and inference efficiency reveal different model rankings.
GPT-6.1 Sol achieves the highest terminal assets, whereas DeepSeek delivers the highest net-asset growth per dollar of estimated API expenditure.
GPT's mean estimated API expenditure is approximately 80 times DeepSeek's, while its net-asset growth is about 2.9 times as large.
GLM-5.3-Flash ranks second on both measures.
DeepSeek also uses more tool calls than GLM while achieving higher API Efficiency, showing that interaction count alone does not capture inference cost.

\begin{figure*}[t]
\centering
\includegraphics[width=\textwidth]{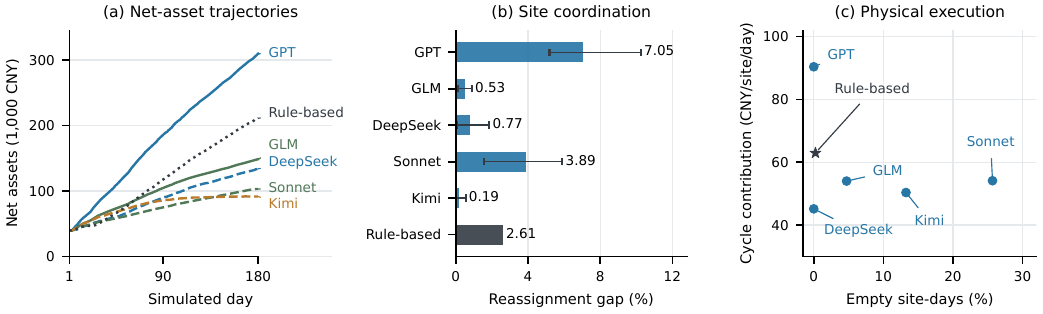}
\caption{Long-horizon operations, site coordination, and physical execution.
(a) Mean net-asset trajectories across three runs per LLM, compared with the rule-based baseline.
(b) Mean reassignment gaps across six checkpoints, with whiskers showing the range across three executions and a fixed baseline.
(c) Availability and visit-cycle returns, averaged across executions.
Contributions exclude final incomplete cycles and costs other than visit fees.}
\label{fig:rq1-3}
\end{figure*}

\section{Analysis}

\subsection{Operational Performance Profiles}
\label{sec:operational-profiles}

To understand the operational strengths and weaknesses underlying the agents' economic outcomes, we analyze their performance across five dimensions in Figure~\ref{fig:operational-profiles}.
No single model leads across all five dimensions, and stronger profitability does not consistently coincide with better inventory control, pricing, or capital efficiency.
We examine these differences below to characterize how each agent combines product selection, inventory management, and resource use during long-horizon operations.

\textbf{Profitability.}
GPT-6.1 Sol achieves the highest mean terminal net assets at 309.35 thousand CNY, followed by GLM-5.3-Flash at 148.43 thousand CNY.
Kimi K3 achieves the lowest return at 91.52 thousand CNY.

\textbf{Assortment.}
GPT-6.1 Sol leads the LLM agents in assortment quality, scoring 0.992.
GLM-5.3-Flash follows at 0.817, while Claude Sonnet 5.5 has the lowest score at 0.500.

\textbf{Inventory.}
Claude Sonnet 5.5 and GLM-5.3-Flash record the lowest waste rates at 0.16\% and 0.27\%, respectively.
DeepSeek V4.1 Flash has the highest rate at 1.85\%.
Inventory availability also differs substantially.
Claude's low waste accompanies empty shelves on 25.21\% of site-days, whereas DeepSeek's machines are empty on only 1.67\%, combining broader inventory availability with greater losses.

\textbf{Pricing.}
Claude Sonnet 5.5 achieves the lowest conditional pricing regret at 1.32 CNY per machine per day, followed by GLM-5.3-Flash at 2.45.
GPT-6.1 Sol has a much larger gap at 30.18, while all other models fall between 1.32 and 4.22.
Its operational profile thus combines the strongest realized profitability with the largest conditional pricing gap.

\textbf{Capital Efficiency.}
Claude Sonnet 5.5 generates the highest daily gross profit per unit of inventory capital at 0.0589, followed by GLM-5.3-Flash at 0.0362.

\subsection{RQ1: Can LLM agents effectively manage long-horizon business operations?}
\label{sec:rq1}

Most evaluated agents can establish profitable operations, but reliable long-horizon outperformance remains limited.
All 15 LLM executions finish above the initial capital, whereas only GPT-6.1 Sol exceeds the rule-based baseline in all three executions.
The trajectories in Figure~\ref{fig:rq1-3}(a) also distinguish sustained growth from early gains followed by stagnation, particularly for Kimi K3.

\subsection{RQ2: Can agents coordinate resources across sites?}
\label{sec:rq2}

Site coordination remains incomplete even for the best-performing agent.
We evaluate this through the best reassignment of existing assortments across sites at six 30-day checkpoints, using the standardization of the assortment metric.
GPT's actual assignment falls 7.05\% below this restricted optimum, compared with 2.61\% for the baseline in Figure~\ref{fig:rq1-3}(b).
Its strong aggregate return therefore coexists with opportunities to improve where its selected products are offered.

Because this diagnostic introduces no new products, the gap identifies unrealized value in matching existing offerings to heterogeneous sites.
Conversely, a small reassignment gap does not establish strong coordination when the underlying assortments are weak.
DeepSeek and Kimi illustrate this distinction through their limited reassignment gains and low assortment scores.
Coordination must therefore be assessed jointly with product quality, since efficient placement cannot compensate for unsuitable offerings.

\subsection{RQ3: Can agents plan physical execution?}
\label{sec:rq3}

Agents do not consistently balance local availability with economical execution.
Over days 16--180, GPT and DeepSeek both avoid completely empty shelves, but GPT achieves approximately twice DeepSeek's visit-cycle contribution per site-day in Figure~\ref{fig:rq1-3}(c).
Sonnet and Kimi instead exhibit persistent empty-site exposure despite holding corresponding products in the warehouse.

These outcomes reflect two separate requirements of physical execution.
Central inventory becomes available to customers only after a paid on-site transfer, so procurement alone cannot maintain local supply.
Conversely, visits that preserve availability can absorb a substantial share of gross profit when they are too costly relative to the sales they support.
Effective execution requires coordinating replenishment timing and visit frequency with local demand.
Cycle returns, measured as sales gross profit minus visit fees divided by cycle duration, must therefore be interpreted alongside availability and the costs excluded from this measure.

\subsection{RQ4: Can agents improve through feedback?}
\label{sec:rq4}

\begin{figure}[t]
\centering
\includegraphics[width=\columnwidth]{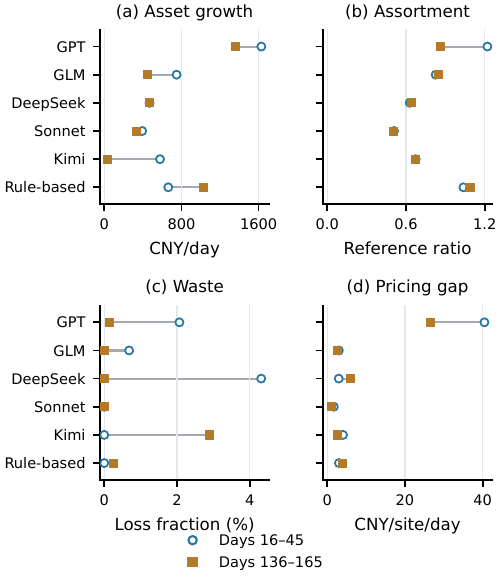}
\caption{Early- and late-period operating outcomes.
Markers show mean LLM outcomes and rule-based baseline outcomes.}
\label{fig:rq4-feedback}
\end{figure}

The trajectories provide limited evidence of consistent economic improvement as feedback accumulates.
Comparing days 16--45 with days 136--165 in Figure~\ref{fig:rq4-feedback}, daily asset growth decreases for all five model means and in 13 of the 15 executions.
Meanwhile, several agents improve individual operating metrics.
Better waste control or pricing therefore does not consistently translate into a stronger overall operating policy.

An accounting decomposition helps explain this disconnect.
DeepSeek generates more gross profit, but additional warehouse and visit expenses offset most of that gain.
These patterns suggest that local corrections can coexist with increasingly costly inventory and execution commitments.
However, changing environmental conditions and inventory states prevent this temporal comparison from causally identifying whether feedback itself improves the agents' decisions.

\subsection{Failure Modes}

Trajectory inspection identifies four observed failures in action feasibility, inventory cost planning, physical execution, and policy revision.
These cases illustrate behavior without establishing prevalence or causal effects.

\textbf{Incomplete Grounding of Action Plans.}
In one Kimi K3 run, all 49 allocations in a day 170 restocking batch were rejected because the requested products lacked sufficient warehouse stock.
The agent had observed 4,701 warehouse units in aggregate and attempted replenishment across seven sites while an order remained in transit.
After rejection, it ended the day and waited for delivery.
The episode shows why allocation plans require product-level availability checks.

\textbf{Insufficient Planning for Inventory Costs.}
In another Kimi run, six purchases between days 136 and 169 repeated an identical bundle of 21 products.
Although core products required replenishment, the bundle also included products with no recorded sales during this interval.
For one such product, 2,158 unexpired units were already in the warehouse before day 169, yet the agent ordered another 80 units.
Procurement thus continued to expand inventory that was not being deployed or consumed.

\textbf{Fragmented Physical Execution.}
In one DeepSeek V4.1 Flash run, a campus machine generated only 5.14 CNY of gross profit between visits on days 125 and 126, against a 35 CNY visit fee.
The agent returned on day 126 to add just four units, incurring another visit fee.
This sequence illustrates how frequent responses to small stock changes can create interventions whose costs are poorly matched to site-level turnover.

\textbf{Ineffective Use of Operating Feedback.}
GLM-5.3-Flash illustrates the difference between immediate error recovery and persistent policy revision.
On day 163, four allocations requesting four units each were rejected because only three units of the product remained in the warehouse.
The agent then successfully allocated those three units to one site.
However, the same four infeasible allocations reappeared on day 168, followed by the same local repair.
The feedback supported correction within each interaction but did not prevent the allocation pattern from recurring.

\section{Conclusion}

We introduced AutoVend, a benchmark for long-horizon autonomous business management across 20 heterogeneous vending sites with shared inventory, procurement lead times, and paid physical execution.
Evaluation of five LLM agents over 180 simulated days shows that four of five underperform a rule-based inventory-control baseline.
Operational diagnostics reveal that strong aggregate returns can coexist with coordination gaps, while improvements in individual metrics need not produce stronger economic growth.
The trajectory cases further expose difficulties in verifying feasible allocations, adapting procurement, consolidating visits, and retaining corrections across decisions.
Future work should test these findings, drawn from three runs per model in one environment, across varied operating conditions and controlled interventions such as explicit state verification and persistent policy updates.

\bibliography{references}

\clearpage
\appendix
\section{Prompts}
\label{app:prompts}

\subsection{System Prompt}
\begin{lstlisting}[style=autovend-prompt]
You operate a network of 20 smart vending machines for 180 days, starting with 40000 in cash.
Each day is one turn: you receive an observation, may call tools, then call end_day.

## Objective
Maximize net assets at the end of the run.
Net assets = cash + the cost value of every unexpired unit you hold, wherever it sits: on a shelf, in the warehouse, or in transit.

- The only thing that increases it is selling above unit cost.
- The things that decrease it: warehouse rent, labor, the daily fee, expired-sale fines, and stock that expires or is discarded unsold.
- Buying stock moves value from cash into inventory; it is not by itself a gain or a loss.

## Starting state
On day 1 every shelf and the warehouse are empty, no order is in transit, and there is no sales history.
The catalog is the only thing that exists: 1372 SKUs.
Nothing sells until you order stock, wait for it to arrive, and restock a device on site.

## Goods flow
    order -> in transit -> warehouse -> restock (on site) -> shelf -> sale

- Ordered units are debited from cash immediately.
- The supplier promises 3 days; actual delivery may be earlier or later.
  Units in transit cannot be shelved and cannot serve a customer.
  Arrivals by day and any delayed orders are in the daily observation.
- Units land in the warehouse.
  The warehouse is the only pool a restock can draw from.
- Reducing or delisting returns units from a shelf to the warehouse; they are not destroyed.
- A unit leaves your system only when a customer buys it or when you discard it.

## Prices
Prices are per (device, SKU), and each of the 20 machines is configured independently.
A price may be set anywhere from 0.50x to 2.00x the SKU's suggested price; out-of-band prices are rejected.

## Costs
Warehouse rent is 0.03 per unit of volume per day.
Every unit of every SKU takes exactly one unit of volume, so rent is charged per unit held and a machine's capacity is a count of units.
On top of rent you pay a fixed fee of 80 per day whether or not you act.

On-site actions (restock, reduce, delist) cost 35 per device per day touched — a device touched any number of times in one day is charged once.
Ordering, repricing, discarding and all queries are remote and cost nothing.
Each tool description says which kind it is.

## Shelf life
Each SKU has a `shelf_life_days` value in the catalog, counted from the day a unit arrives in the warehouse.
Restocking, reducing and delisting do not reset that clock.
Expired units do not disappear, and they are worth nothing in net assets.

- Expired units on a shelf still sell, and every expired unit sold costs an extra 1000.
  Delisting sends them back to the warehouse, which stops the fine exposure but does not make them sellable again.
- Expired units in the warehouse keep taking up volume and incurring rent.
  They cannot be restocked into anything but more expired stock; discarding is the only way to remove them.

## Funding
Cash can go negative.
Rent, the daily fee, visits and purchases keep being charged, no interest is added, and the run does not end.
A negative balance counts against net assets one for one, so buying on credit does not change net assets by itself — it moves cash into inventory at cost.
What it does cost is rent for every day the stock sits, and 3 days before it can be sold.

## Environment
Shanghai, CN (Asia/Shanghai).
All amounts are in CNY.
Temperature, rain and public holidays affect demand, differently by product category and by site.

## Demand
- Fill level: a fuller device sells faster.
  Free volume depresses demand for every item on that device, not only for the missing ones.
- Depth: an item's sales stop increasing beyond a few units on the shelf; further units only accrue rent.
- Noise: daily sales are random around that item's underlying rate, and that rate is what a longer window of observations estimates.
- Demand rate, price response and product fit all vary with location type.
  A product that sells at one site can be dead stock at another, and the same price change can help at one site and hurt at another.

## Memory
Saved notes are never trimmed or shortened.
They are limited to 6,000 characters.

## Operating loop
There is no human user: no one will prompt you, approve a plan, or flag a problem.
When the day's work is done, call end_day.
\end{lstlisting}

\subsection{User Prompt}
\textbf{Daily user prompt template.}
\begin{lstlisting}[style=autovend-prompt]
Day {day} of {horizon_days} — {weekday}, {date}.

{observation}

Keep running the business to maximize your net assets.
When today's work is done, call end_day.
\end{lstlisting}

\textbf{Initial daily prompt example.}
The initial daily prompt example is generated from the simulator's starting state rather than taken from a historical agent run.
\begin{lstlisting}[style=autovend-prompt]
Day 1 of 180 — Wednesday, 2025-01-01.

{
 "balance": {
  "cash":39920.0,
  "shelf":0.0,
  "warehouse":0,
  "in_transit":0.0,
  "inventory_at_cost":0.0,
  "net_assets":39920.0
 },
 "pnl": {
  "columns":[
   "window","days","units_sold","revenue","gross_profit",
   "rent","labor","daily_fee","expired_fine","net"
  ],
  "rows":[
   ["yesterday",0,0,0.0,0.0,0.0,0.0,0.0,0.0,0.0],
   ["last_7d",0,0,0.0,0.0,0.0,0.0,0.0,0.0,0.0],
   ["last_30d",0,0,0.0,0.0,0.0,0.0,0.0,0.0,0.0]
  ]
 },
 "devices": {
  "columns":[
   "device_id","location","capacity","units_on_shelf",
   "n_skus","empty_slots","sold_yesterday","sold_7d_avg",
   "expired_units","expiring_7d_units","days_since_visit"
  ],
  "rows":[
   ["CN_WORKSITE_01","worksite/canteen",200,0,0,0,0,0.0,0,0,null],
   ["CN_WORKSITE_02","worksite/canteen",200,0,0,0,0,0.0,0,0,null],
   ["CN_WORKSITE_03","worksite/canteen",200,0,0,0,0,0.0,0,0,null],
   ["CN_WORKSITE_04","worksite/plant_floor",200,0,0,0,0,0.0,0,0,null],
   ["CN_WORKSITE_05","worksite/plant_floor",200,0,0,0,0,0.0,0,0,null],
   ["CN_WORKSITE_06","worksite/plant_floor",200,0,0,0,0,0.0,0,0,null],
   ["CN_CAMPUS_01","campus/university",200,0,0,0,0,0.0,0,0,null],
   ["CN_CAMPUS_02","campus/school_dormitory",200,0,0,0,0,0.0,0,0,null],
   ["CN_RESIDENTIAL_01","residential/apartment",200,0,0,0,0,0.0,0,0,null],
   ["CN_RESIDENTIAL_02","residential/apartment",200,0,0,0,0,0.0,0,0,null],
   ["CN_RESIDENTIAL_03","residential/apartment",200,0,0,0,0,0.0,0,0,null],
   ["CN_RESIDENTIAL_04","residential/dormitory",200,0,0,0,0,0.0,0,0,null],
   ["CN_COMMERCIAL_01","commercial/sports_venue",200,0,0,0,0,0.0,0,0,null],
   ["CN_COMMERCIAL_02","commercial/hotel",200,0,0,0,0,0.0,0,0,null],
   ["CN_COMMERCIAL_03","commercial/hotel",200,0,0,0,0,0.0,0,0,null],
   ["CN_COMMERCIAL_04","commercial/sports_venue",200,0,0,0,0,0.0,0,0,null],
   ["CN_COMMERCIAL_05","commercial/leisure_venue",200,0,0,0,0,0.0,0,0,null],
   ["CN_COMMERCIAL_06","commercial/hotel",200,0,0,0,0,0.0,0,0,null],
   ["CN_OFFICE_01","office/industrial_park",200,0,0,0,0,0.0,0,0,null],
   ["CN_OFFICE_02","office/office_building",200,0,0,0,0,0.0,0,0,null]
  ]
 },
 "warehouse": {
  "units":0,
  "n_skus":0,
  "value_at_cost":0,
  "rent_per_day":0.0,
  "expired_units":0,
  "expiring_7d_units":0
 },
 "in_transit": {
  "orders":0,
  "units":0,
  "value_at_cost":0.0,
  "arrivals":[],
  "delayed":[]
 },
 "calendar": {
  "today":{"is_workday":false,"holiday_name":"\u5143\u65e6"},
  "next_14d":[]
 }
}

Keep running the business to maximize your net assets.
When today's work is done, call end_day.
\end{lstlisting}

\section{Tools}
\label{app:tools}

\begin{table*}[t]
\centering
\caption{Tools available to the AutoVend agent.
A question mark denotes an optional parameter.}
\label{tab:tools}
\small
\setlength{\tabcolsep}{4pt}
\begin{tabular}{@{}p{0.27\textwidth}p{0.27\textwidth}p{0.42\textwidth}@{}}
\toprule
\textbf{Tool} & \textbf{Parameters} & \textbf{Function and constraints} \\
\midrule
\multicolumn{3}{@{}l}{\textbf{Remote operational tools}} \\
\texttt{place\_order} & \texttt{items[\{sku\_id, qty, supplier\_id?\}]} & Order warehouse stock and debit cash immediately.
The default supplier is \texttt{default}. \\
\texttt{adjust\_price} & \texttt{items[\{device\_id, sku\_id, price\}]} & Set a machine-specific price within 0.5--2.0 times the reference price.
The product need not already be stocked. \\
\texttt{discard} & \texttt{items[\{sku\_id, qty\}]} & Destroy warehouse units, taking the earliest-expiring units first. \\
\texttt{end\_day} & None & Close the daily action window. \\
\texttt{get\_catalog} & \texttt{query?}, \texttt{category\_l1?}, \texttt{category\_l2?}, \texttt{price\_min?}, \texttt{price\_max?}, \texttt{in\_warehouse\_only?}, \texttt{sort\_by?}, \texttt{page?}, \texttt{page\_size?} & Search public product attributes and warehouse availability.
Sorting uses arbitrary catalog order, ascending or descending reference price, or reference margin. \\
\texttt{get\_categories} & \texttt{category\_l1?} & Return subcategory counts, reference price ranges, and median reference margins. \\
\texttt{get\_warehouse} & \texttt{page?}, \texttt{page\_size?} & Return warehouse quantities, costs, and expiry information by product. \\
\texttt{get\_device\_detail} & \texttt{device\_id}, \texttt{window\_days?} & Return product-level shelf quantities, prices, daily sales, and expiry information for one machine. \\
\texttt{get\_sku\_across\_devices} & \texttt{sku\_id}, \texttt{window\_days?} & Return a product's stock and realized sales across machines, together with site types. \\
\texttt{get\_sales\_history} & \texttt{window\_days?} & Return daily network sales, operating costs, visits, and the top 20 products by realized units sold in the window. \\
\texttt{query\_sql} & \texttt{sql} & Execute a read-only SQLite query over operating records.
Return at most 500 rows and stop queries exceeding two seconds. \\
\texttt{get\_calendar} & \texttt{days\_ahead?} & Return already published holidays and makeup workdays.
The default horizon is 60 days, with a range of 1--400. \\
\texttt{get\_weather} & None & Return settlement weather for today and the following seven days. \\
\midrule
\multicolumn{3}{@{}l}{\textbf{Physical execution tools}} \\
\texttt{restock} & \texttt{items[\{device\_id, sku\_id, qty\}]} & Transfer warehouse units to a machine, subject to stock availability and shelf capacity. \\
\texttt{reduce} & \texttt{items[\{device\_id, sku\_id, qty\}]} & Return a specified shelf quantity to the warehouse, taking the earliest-expiring units first. \\
\texttt{delist} & \texttt{items[\{device\_id, sku\_id\}]} & Return all remaining shelf units of a product to the warehouse. \\
\midrule
\multicolumn{3}{@{}l}{\textbf{Memory tools}} \\
\texttt{read\_memory} & None & Read the persistent notebook in full. \\
\texttt{write\_memory} & \texttt{text} & Replace the notebook with at most 6,000 characters.
An empty string clears it. \\
\bottomrule
\end{tabular}
\end{table*}

A question mark denotes an optional parameter.
The six inventory and pricing mutation tools accept batches of 1--100 items, with quantities specified as positive integers.
Structurally malformed batches are rejected before execution.
For a structurally valid batch, operational constraints are checked separately for each item, so valid items execute even when other items are rejected.
Results report accepted and rejected items together with rejection reasons.
Rejected items do not change inventory or trigger a visit.
The three physical tools incur one visit charge per machine per day, regardless of the number of successful operations.
Query windows are restricted to 1, 3, 7, or 30 days, with a default of 30 days.
Paginated queries default to 20 rows per page and allow at most 50.
A notebook write exceeding the character limit is rejected without replacing the existing notes.

\section{Ethical Use of Data}
\label{app:ethics}
The order-level sales records used to build the simulator were provided by a commercial smart vending company with its authorization.
The records contain no personal information about customers or staff.
We release only the calibrated simulator parameters, not the raw records.

\end{document}